\documentclass{article}
\usepackage{arxiv}
\usepackage{sectsty}
\usepackage{subcaption}
\usepackage{titlesec}
\usepackage{graphicx}
\usepackage{todonotes}
\usepackage{listings}

\usepackage{comment}
\usepackage{amsmath}
\usepackage{csquotes}
\usepackage[backend=biber,sorting=nyt,bibencoding=latin1,style=numeric,maxnames=2,minnames=1]{biblatex}

\usepackage[stable]{footmisc}

\usepackage[utf8]{inputenc}
\usepackage[T1]{fontenc}

\usepackage{hyperref}
\usepackage{url}

\titlespacing*{\section}{0pt}{8pt}{8pt}

\hypersetup{colorlinks=true, urlcolor=blue, linkcolor=blue, filecolor=blue, citecolor=blue, breaklinks=true, 
}
\usepackage{footmisc}
\begin{document}

\title{Extraction of Common Conceptual Components \\ from Multiple Ontologies}

\author{Luigi Asprino\\
Department of Classical Philology and Italian Studies \\
University of Bologna \\
Via Zamboni 32, 40126 Bologna, Italy \\
\texttt{luigi.asprino@unibo.it}
\And
Valentina Anita Carriero\\
Department of Computer Science and Engineering \\
University of Bologna \\
Mura Anteo Zamboni 7, 40126 Bologna, Italy \\ 
\texttt{valentina.carriero3@unibo.it} 
\And
Valentina Presutti \\
Department of Modern Languages, Literatures, and Culture \\
University of Bologna \\
Via Cartoleria 5, 40124 Bologna, Italy \\ 
\texttt{valentina.presutti@unibo.it}
}

\maketitle

\begin{abstract}
Understanding large ontologies is still an issue, and has an impact on many ontology engineering tasks.
We describe a novel method for identifying and extracting conceptual components from domain ontologies, which are used to understand and compare them. The method is applied to two corpora of ontologies in the Cultural Heritage and Conference domain, respectively. The results, which show good quality, are evaluated by manual inspection and by correlation with datasets and tool performance from the ontology alignment evaluation initiative.
\end{abstract}




\maketitle

\section{Introduction}
\label{sec:intro}
\begin{figure}[t] \centering
\includegraphics[width=0.7\linewidth]{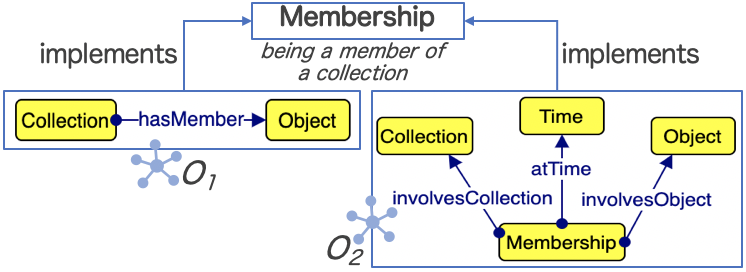}
	\caption{Implementations of the Membership CC from two different ontologies.}
	\label{img:membership}
\end{figure}

Understanding large ontologies - by humans or machines - is both a struggle and crucially important for performing ontology engineering tasks such as ontology reuse, ontology matching, ontology evaluation, and (federated) querying \cite{Carriero2020reuse}. 
According to \cite{DBLP:journals/ker/DudasLSP18}, existing visualisation tools fail in providing overviews of large ontologies, which is crucial for ontology understanding, while none of them allows to compare multiple ontologies. Besides the layout and interaction features, the problem lays in the lack of effective methods for producing summaries of large ontologies.
Many summarisation approaches focus on analysing the data level, e.g. to reduce the size of a knowledge graph and allow simplified queries for testing its coverage \cite{summarization-survey2018-arxiv,cebiric2018summarizing}.
Available summarisation methods addressing the conceptual level are based on \textit{extractive} approaches that select and return a subset of nodes from the original ontology, i.e. the key concepts, as a summary \cite{summarization-survey2018-arxiv}. However, an overall understanding of all the \emph{facts} an ontology can represent, and a comparison between multiple ontologies, are not supported.
For example, we may identify that in a cultural heritage ontology the concepts \textit{Cultural Property} and \textit{Collection} are key ones, however this is insufficient to understand if one ontology allows to answer whether a cultural property has been in a collection. Two ontologies having the same key concept would appear they address the same modelling problem, which may not be the case.
For instance, an ontology $O_{1}$ may implement a \emph{membership} relation between an object and a collection, i.e. \enquote{being a member of a collection}, as an object property \texttt{hasMember} between a class \texttt{Collection} and a class \texttt{Object}, while an ontology $O_{2}$ may implement it as an n-ary relation class \texttt{Membership} connected to three arguments e.g., the classes \texttt{Collection}, \texttt{Time}, and \texttt{Object}\footnote{Other implementations are possible.} (Figure \ref{img:membership}). We refer to these implementations as observed ontology design patterns \cite{shortGangemi2005}, intended as adopted modelling solutions that can be observed in existing ontologies, regardless their correctness or quality level, which may or may not reuse reference ontology design patterns (ODPs).

A \emph{conceptual component} (CC) is a complex (cognitive) relational structure that a designer implements in an ontology by using classes, properties, axioms, etc. Examples of CCs are  membership, locating, interpreting, observing. 
This notion of CC is inspired by the concept of \textit{knowledge pattern} presented in \cite{DBLP:journals/semweb/GangemiP10}. 
Conceptual components are cognitive objects: they are the \textit{intensional} counterparts of OWL implementations in semantic web ontologies\footnote{OWL has purely extensional semantics.}. A CC may be \emph{implemented} by means of different ontology fragments, the observed ontology design patterns (OODPs), across different ontologies. Therefore, the CCs emerging from an ontology (corpus) indicate which types of facts, rather than which types of entities, an ontology (corpus) can represent. The OODPs implementing a CC show the adopted modelling solutions by a designer: which competency questions \cite{Gruninger1995} and inferences are supported by an ontology.\\
Our approach aims at identifying the CCs implemented in multiple ontologies, to support their understanding and comparison. We group OODPs from different ontologies in semantically-meaningful clusters, i.e. CCs. These clusters provide a conceptual ordering, based on CC, over the different implementations (OODPs), hence providing a means to e.g. identify the most appropriate OODP to reuse or align, based on specific requirements.
While addressing this issue is an interesting research result \emph{per se}, the method we propose may lead to novel approaches to ontology engineering tasks such as pattern-based ontology reuse, ontology visualisation, ontology matching, ontology evaluation. This method is also relevant from an empirical perspective on knowledge engineering, that is to observe the common conceptual issues and modelling solutions adopted in ontologies, with potentially a strong impact on semantic web interoperability. \\
\indent In contrast to existing methods, we develop a \emph{non-extractive} technique, as the identified conceptual components are not part of the original sources. To the best of our knowledge, this is the first approach of this kind. Our implementation combines community detection, word sense disambiguation, frame detection and clustering techniques. By applying it on a corpus of ontologies from a knowledge domain, it produces a \textit{catalogue of CCs and their corresponding observed ontology design patterns} organised as a hierarchical network. The CCs are labeled and linked to their OODPs from the corpus. Therefore, the ontologies are classified based on the CCs that they implement.
%
We apply our method to a corpus of 43 cultural heritage ontologies and to a corpus of 16 Conference ontologies used in the ontology alignment evaluation initiative (OAEI)\footnote{\label{note:oaei}\url{http://oaei.ontologymatching.org/}}. All software, input data, and results are available online\footnote{\label{note:package}\url{https://github.com/stlab-istc-cnr/conceptual-components}} as a GitHub repository. We evaluate our results using two approaches: 1) manual inspection of the resulting OODPs and conceptual components, and 2) correlation of our results with ontology alignment tools and datasets from the ontology alignment evaluation initiative\footref{note:oaei}.

The contribution of this paper can be summarised as follows: (i) we define a novel method for multiple ontology summarisation, based on conceptual components and observed ontology design patterns; (ii) we implement the proposed method as a non-extractive technique; (iii) we produce and publish a catalogue of conceptual components and observed ODPs from a corpus of CH ontologies.

Section \ref{sec:material} describes the datasets used as input sources. Section \ref{sec:method} illustrates our approach and its implementation. Section \ref{sec:results} reports our experiments and results, while Section \ref{sec:discussion} focuses on their evaluation and discussion. Before concluding in Section \ref{sec:conclusion}, we discuss related works in Section \ref{sec:related}.
\section{Input source}
\label{sec:material}
Our empirical basis is composed of two ontology corpora. 

\noindent \textbf{Cultural Heritage.} We build a corpus consisting of 43 Cultural Heritage (CH) ontologies\footref{note:package}. The motivation for choosing this domain is twofold: (i) we have experience in modelling CH ontologies, and (ii) the requirements of CH ontologies are generally complex, hence we hypothesise that it provides a good testbed for the generalisability of our method.
Ontologies that focus on related domains (e.g. geometry, chemistry) and top-level ontologies have been excluded.
To select the ontologies we used two main sources.
We analysed and searched the \emph{Vocabs} section of the Linked Open Vocabularies Repository (LOV)\footnote{\url{https://lov.linkeddata.es}} by filtering the results using tags related to CH (such as Catalogs, FRBR, Metadata).
Moreover, we disseminated a call to fill a Survey\footnote{\url{https://t.co/ghwk6lxCOH?amp=1}} on 3 CH- and ontology engineering-related mailing lists and on social networks: 40 people, mostly researchers, participated in the survey. People were asked to (i) indicate which ontologies they already knew, from the list of ontologies selected from LOV; (ii) recommend other ontologies;
(iii) indicate in which projects they had used any of them.
Almost all ontologies were known by at least one participant,
and four CH ontologies have been recommended and added to the corpus.
For each ontology, the latest version available is included in the corpus. Four of them are ontology networks i.e. composed of multiple modules:
we consider each networked ontology as one ontology. 
When possible, we include the inferred version (i.e. with materialised inferences) of the ontologies. To this end we use the OWL API\footnote{\url{https://github.com/owlcs/owlapi}} and the HermiT Reasoner\footnote{\url{http://www.hermit-reasoner.com}}. 
Due to import- and inconsistency-related problems, for 10 ontologies 
we only include the asserted version.\\
The resulting CH corpus (cf. Table \ref{tab:input}) counts a total number of 2,707 classes (\texttt{owl:Class}, \texttt{rdfs:Class}), with an average of $\sim$63 classes per ontology.
As for the properties (\texttt{owl:ObjectProperty}, \texttt{owl:Datatype\allowbreak{}Property}, \texttt{rdf:Property}), they are 9,132 in total, with an average of $\sim$212 per ontology
The total number of logical axioms is 26,392, with an average of $\sim$613 per ontology

\noindent \textbf{Conference.} Our second corpus is provided by the dataset of the Conference evaluation track\footnote{\url{http://oaei.ontologymatching.org/2020/results/conference}} of the OAEI 2020 campaign (\emph{Conf} for short), which contains 16 ontologies\footnote{\url{https://owl.vse.cz/ontofarm/\#ontologies}} on a specific domain, less vast than CH but with a good range of subtopics and related domains (e.g. price, travel). 
This corpus counts (cf. Table \ref{tab:input}) a total number of 851 classes, with an average of $\sim$53 classes per ontology;
the total number of properties 
is 714, with an average of $\sim$44 properties per ontology.
For all 16 ontologies the inferred versions have been computed. The total number of logical axioms is 4,097, with an average of $\sim$256 axioms per ontology.

\begin{table*}[t]
\begin{center}
	\caption{Corpora of ontologies: statistics}
	\label{tab:input}
	\resizebox{\textwidth}{!}{ 
	\begin{tabular}{|p{1.5cm}|c|cccc|cccc|cccc|}
	\hline 
		{ \bf Dataset } & {\bf\# ontologies} & \multicolumn{4}{c|}{{\bf \# logical axioms}} & \multicolumn{4}{c|}{{\bf \# classes}} & \multicolumn{4}{c|}{{\bf \# properties}} \\\hline 
		
		& &tot & avg & min & max & tot & avg & min & max  & tot & avg & min & max \\
		\emph{CH}  & 43 & 26,392 & $\sim$613 & 16 & 1,060 & 2707 & $\sim$63 & 5 & 539 & 9132 & $\sim$212 & 6 & 4324  \\
		
		\emph{Conf}  & 16 & 4097 & $\sim$256 & 65 & 739 & 851 & $\sim$53 & 14 & 140 & 714 & $\sim$44 & 17 & 78  \\
		
		\hline
	\end{tabular}
	}
	\end{center}
\end{table*}
\section{Approach}
\label{sec:method}
\begin{figure}[t] \centering
\includegraphics[width=0.9\linewidth]{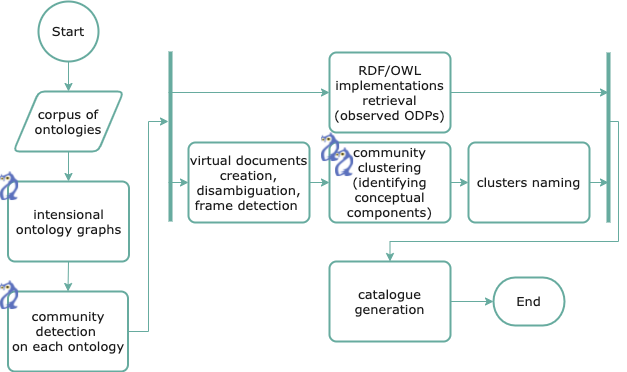}
	\caption{Approach for conceptual components extraction. One owl-logo means that the process works on one ontology at a time, two owl-logos that it works on the whole corpus.}
	\label{img:pipeline}
\end{figure}

The intuition (and assumption) behind our method (summarised in Figure \ref{img:pipeline}) is that ontologies are designed (either intentionally or unintentionally) as compositions of conceptual components, implemented by (observed) ODPs (an adopted modelling solution).
An ODP\footnote{In this context, by ODP we refer to the notion of \emph{Content ODP}.} captures some relational meaning e.g. membership, observation, participation. We hypothesise that OODPs emerge because 1) their vocabulary is semantically coherent with the relation they represent, i.e. the combination of terms of an OODP evokes that relation. For example, in an OODP \emph{Membership} a possible vocabulary may include the terms: collection, is member of, has member, has unifying property; 2) the density of their internal connections is higher than the density of the connections between them. For example, consider an ontology that models the address of an object as a class \texttt{Address} having four arguments: the object, the city, the street and number, and the postal code. They form an OODP \textit{Address}. Now consider that the same ontology also includes an OODP \textit{Event}, modelling events and their participants. The connections between the entities involved in \textit{Address}, and the connections between the entities involved in \textit{Event} will be denser than the connections between \textit{Address} and \textit{Event}. Community detection algorithms, such as \cite{clauset2004finding}, are able to recognise this topological phenomenon. \\
Our method, depicted in Figure \ref{img:pipeline},
detects the communities in each ontology from a corpus (cf. Section \ref{sec:community}), after a pre-processing step named \textit{intensional ontology graphs} (cf. Section \ref{sec:ig}). Each community potentially identifies an OODP. Then, we retrieve the OWL/RDF\footnote{OWL ontologies, RDF vocabularies.} fragments corresponding to all communities (the actual OODPs) and store them for later use. At the same time, each community is associated with a virtual document: a bag of words generated by concatenating the vocabulary terms describing its entities (e.g. rdfs:label). After a disambiguation and a frame detection steps performed on these virtual documents, they are submitted to a clustering algorithm (cf. Section \ref{sec:clustering}). As a result we obtain a set of clusters, each grouping communities from different ontologies. Based on our assumptions, each cluster is a manifestation of a conceptual component, and each OODP is one of its possible OWL/RDF implementations. We use some heuristics for naming the clusters, and finally we generate a \emph{catalogue}, which provides an abstract, indexed summary of the whole ontology corpus. 

%

\subsection{Intensional ontology graph}
\label{sec:ig}
Most community detection algorithms manipulate undirected graphs and ignore labels: they focus on the topological structure of a network. Therefore, we need to transform our ontologies into graph structures that can be processed by these algorithms, while preserving as much as possible the information about how the ontologies formalise their conceptualisations.
To this aim, we introduce the concept of \emph{intensional ontology graph}, which is a graph derived from an ontology where the nodes represent its predicates (both classes and properties) and the arcs indicate that there is a meaningful relation between two predicates. Informally, this graph encodes the intensional level of the ontology.
Formally, we transform an ontology to its intensional graph by applying the rules defined in Listing \ref{listing:sparql-int}. With the notation \textit{edge\_label\allowbreak{}(source\_label, target\_label)}, we indicate a pair of nodes \textit{source\_label, target\_label} that are connected by the arc \textit{edge\_label}, in the intensional graph. To indicate undirected and unlabelled arcs we use \textit{no\_label}. A rule is a set of premises, expressed in turtle syntax, and a conclusion, expressed with the introduced notation, that follows the symbol \enquote{$\to$}.
\begin{lstlisting}[basicstyle=\normalsize, frame=single, caption={Transformation rules from an OWL/RDF ontology to its corresponding intensional graph.}, label={listing:sparql-int},mathescape]
(r1) :p rdfs:domain :d  .  :p rdfs:range :r . $\to$  :p(:d, :r)

(r2) :c1 rdfs:subClassOf | owl:equivalentClass [
  owl:onProperty :p ; 
  owl:someValuesFrom | owl:allValuesFrom | owl:hasValue |
  owl:maxCardinality | owl:minCardinality | owl:cardinality :c2 ] 
    $\to$  :p(:c1, :c2)
    
(r3) :p(:n1, :n2) $\to$ $no\!\!-\!\!label$(:n1, :n1-p-n2) $no\!\!-\!\!label$(:n1-p-n2, :n2)
\end{lstlisting}
Given an ontology $O$, the first rule (r1) generates an arc :$p$ connecting two nodes, :$d$ and :$r$, for all properties that have domain :$d$ and range :$r$. We ignore domain/range declarations involving blank nodes. Properties without domain/range declarations are assumed to have \texttt{owl:Thing} as domain/range.
Property restrictions (existential, universal, cardinality) generate an edge :$p$ between the class local to the restriction and the class in the restriction expression (r2). We ignore all class expressions, that is we only consider named classes or datatypes. While this may cause some loss of information, we empirically verified on our corpora that the impact is almost insignificant: only 1.62\% of subClassOf/equivalence axioms and 5.42\% of domain/range axioms involve class expressions in the CH corpus, while 1.48\% of subClassOf/equivalence axioms and 9.22\% of domain/range axioms involve class expressions in the Conf dataset.

Class hierarchy and equivalence relations between named classes are left off the intensional graph, to avoid merely taxonomic patterns,
but they are reintroduced when the OWL/RDF OODPs are retrieved (cf. Subsec. \ref{sec:community}).

Rules (r1) and (r2) produce a labelled \textit{multi-graph} (a graph having multiple edges). The last rule (r3) transforms the resulting intensional graph to a corresponding unlabeled and undirected graph structure. For each arc :$p$ between two nodes :$n1$ and :$n2$ it generates two unlabelled arcs. The first connecting $n1$ to a new node :$n1-p-n2$, the second connecting :$n1-p-n2$ to :$n2$. The node :$n1-p-n2$ captures the meaning of the property :$p$, contextualised to its use for connecting :$n1$ and :$n2$. This is a crucial detail for maximising the quality of the detected communities. In fact, communities are disjoint, hence if we only store the information of a property :$p$, this property will only fall into one community. Nevertheless, a same property :$p$ may be relevant in different contexts (and OODPs) and these contexts are captured by its local usage, i.e. the predicates it connects. With this representation we enable overlapping communities, which is crucial to capture concepts that are relevant to more than one pattern.
\begin{figure}[h]\centering
\begin{subfigure}{\textwidth}\centering

  \includegraphics[width=0.95\textwidth]{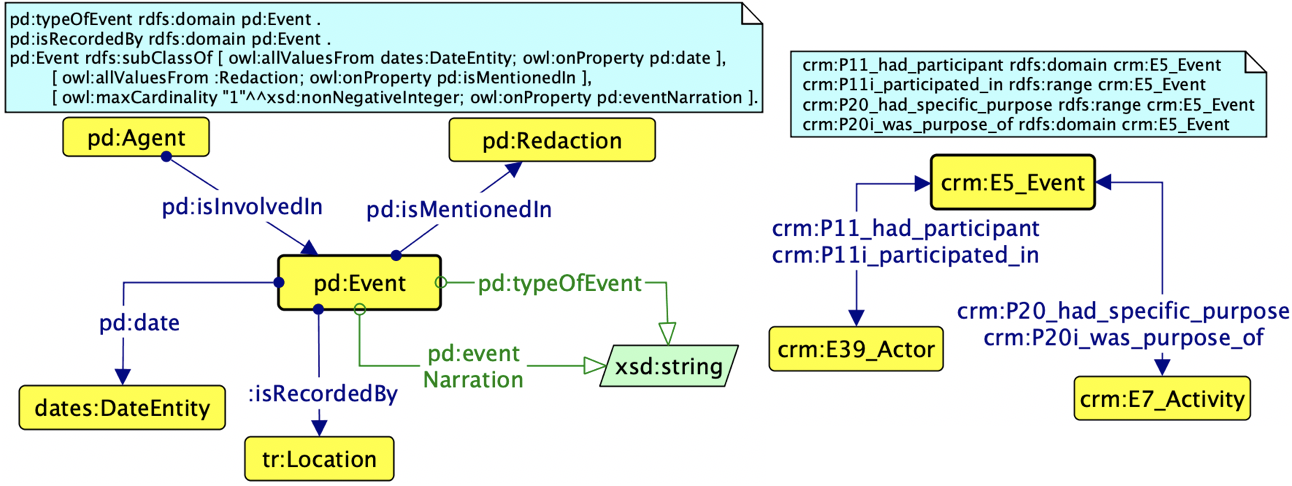}  
  \caption{Two OODPs from the CH corpus.}
  \label{img:owl}
\end{subfigure}
\begin{subfigure}{\textwidth}
  \includegraphics[width=0.95\textwidth]{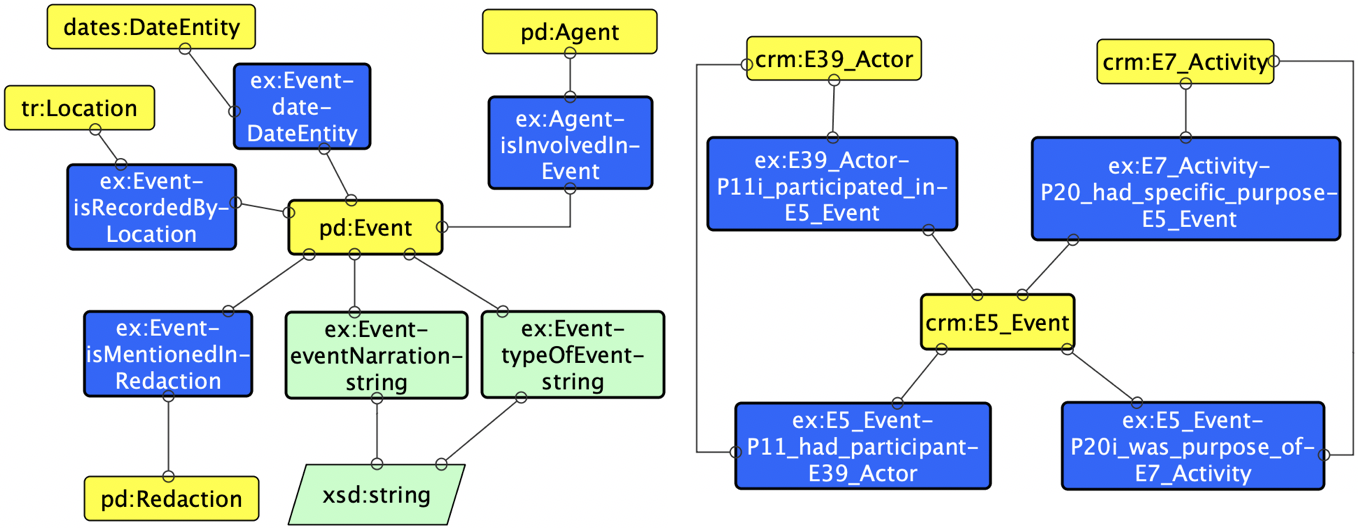}  
  \caption{Intensional graphs corresponding to the OWL OODPs in \ref{img:owl}.}
  \label{img:intensional}
\end{subfigure}
\caption{Example of OWL OODPs and their corresponding intensional ontology graphs. Blue rectangles indicate object properties, green rectangles data properties.}
\label{img:owl-to-intensional}
\end{figure}
We transform each ontology from the two corpora into its \emph{intensional} graph. Figure \ref{img:owl-to-intensional}\footnote{\texttt{pd:} \url{http://postdata.linhd.uned.es/ontology/postdata-core\#} \texttt{tr:} \url{http://postdata.linhd.uned.es/ontology/postdata-transmission\#} \texttt{dates:} \url{http://postdata.linhd.uned.es/ontology/postdata-dates\#} \texttt{crm:} \url{http://www.cidoc-crm.org/cidoc-crm/}} shows two OODPs (in \ref{img:owl}) from POSTDATA (on the left) and CIDOC CRM (on the right) and their corresponding intensional graphs (in \ref{img:intensional})\footnote{We use the \href{https://essepuntato.it/graffoo/}{Graffoo} diagram notation.}.


\subsection{Community detection}
\label{sec:community}
\begin{figure*}[t] \centering
\includegraphics[width=0.85\textwidth]{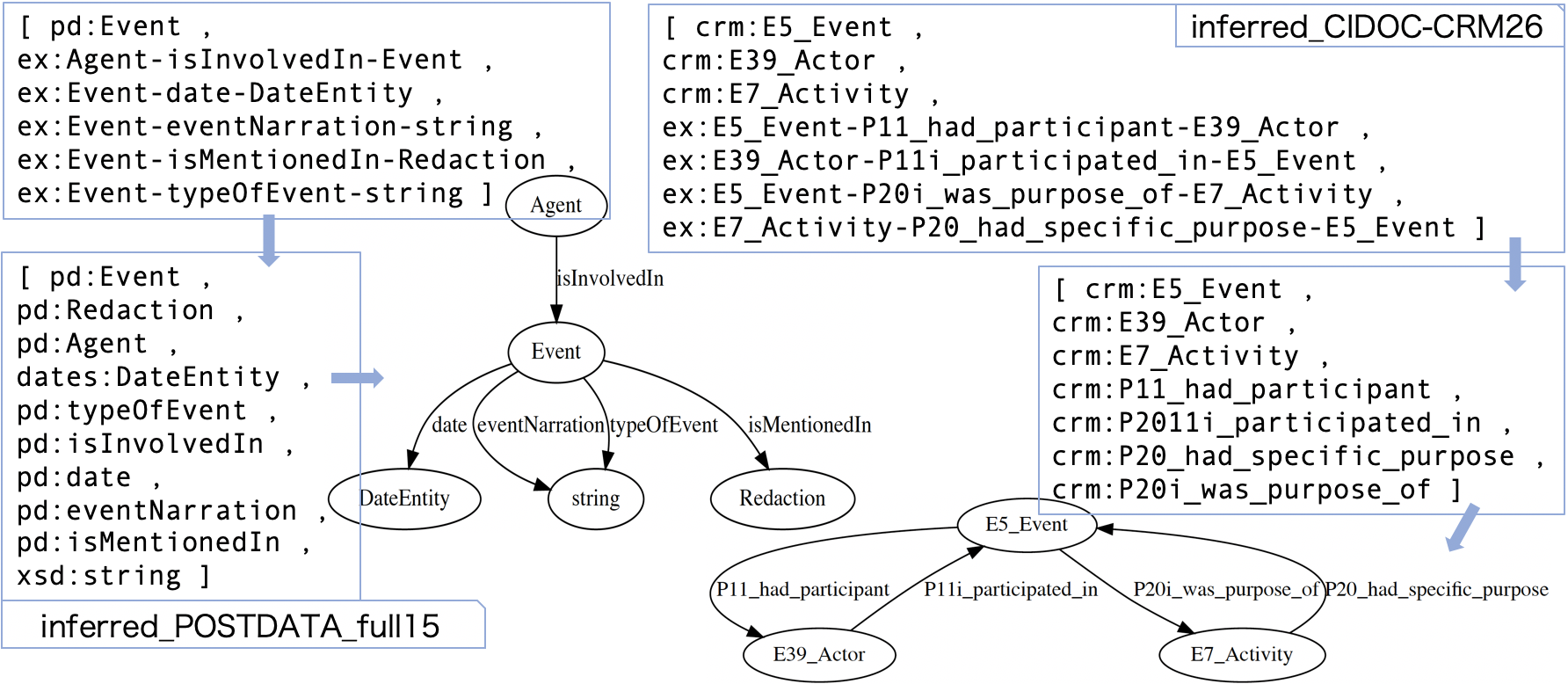}
	\caption{Example of communities detected from two ontologies of the CH corpus.}
	\label{img:communities}
\end{figure*}

Community detection aims at gathering the vertices of a network into groups, such that there is a higher density of edges within groups than between them. 
For detecting the community structure of each ontology, we use the Clauset-Newman-Moore algorithm \cite{clauset2004finding}. This algorithm is based on the greedy optimization of the \emph{modularity}, i.e. a measure of how much the computed division is good in terms of the ratio between the number of edges inside the communities and the number of edges between them. Initially, there are as many communities as the vertices, with each vertex being the only member of its own community, then the two communities that, if merged, most increase the modularity, are repeatedly joined, until it is no longer feasible to merge communities without decreasing the modularity.
%
After running this algorithm on the intensional ontology graphs, we observe that the detected communities highly differ in their density, and that communities with lower density could be further split into meaningful subcommunities. After running some experiments, we found that recursively running the algorithm on communities with density lower than the average density of all communities detected at the previous step, would improve the results. Therefore, the algorithm has been modified to behave in this way (until there is no community that can be split further). \\
\textbf{OWL/RDF OODPs retrieval.} Communities are represented as sets of nodes. In order to further investigate their structure and content, we retrieve the OWL/RDF ontology fragments that contain the original nodes (classes and properties): the observed ODPs. To define their boundary, we use the following heuristics:
for each node in the community, we retrieve the triples asserting its type. As for properties, we retrieve domain and range axioms, inverse, super- and equivalent properties. We retrieve all super- and equivalent classes, and all restrictions that involve at least one property within the community.
Figure \ref{img:communities} shows the sets of nodes retrieved from the two communities depicted in Figure \ref{img:intensional} (from POSTDATA and CIDOC)\footnote{Arrows mean consecutive steps.}. 
They almost perfectly overlap with the ontology fragments in Figure \ref{img:owl}.

\subsection{Clustering and Catalogue Generation}
\label{sec:clustering}
Communities are recognised based on the intensional graph's topological features. Our hypothesis is that they identify OODPs, hence the terms in their vocabulary shall
concur to evoke the relational meaning captured by these OODPs. These relational meanings correspond to (possible specialisations of) the \emph{conceptual components} that we are looking for. As we are working with multiple ontologies, if we cluster the communities according to their vocabularies, we may identify CCs that are shared by (potentially) all of them.  

\begin{figure*}[t] \centering
\includegraphics[width=0.85\textwidth]{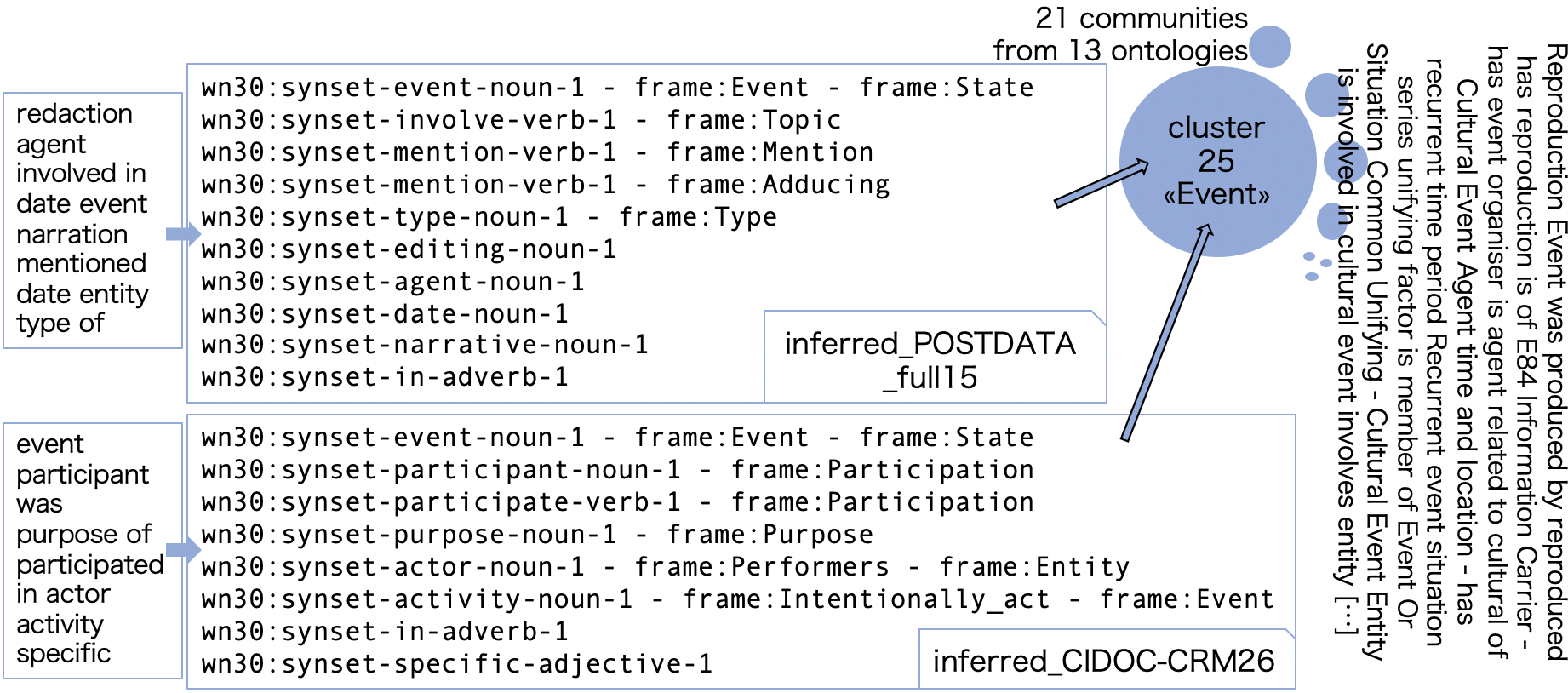}
	\caption{Virtual document disambiguation, frame detection and clustering on the communities from Fig. \ref{img:communities}.}
	\label{img:synsetframe-cluster}
\end{figure*}

\noindent\textbf{Clustering input.} We build a virtual document for each community by concatenating all \texttt{rdfs:label} values from  its entities. We take all English terms and, when no label is present, we use local IDs.
We remove all repetitions and exclude comments, since they may introduce noise. 
Entities with namespaces \texttt{owl:}, \texttt{rdf:}, \texttt{rdfs:} and \texttt{xsd:} are excluded.
%
%
We disambiguate all virtual documents by using UKB\footnote{\url{https://github.com/asoroa/ukb}}, which is based on WordNet (English) version 3.0\footnote{\url{https://wordnet.princeton.edu/}}. Then, we query the profile \textit{B} of Framester\footnote{\url{https://github.com/framester/Framester}}, a knowledge graph that connects many linguistic resources (including WordNet and FrameNet\footnote{\url{https://framenet.icsi.berkeley.edu/fndrupal/}}), for extracting all FrameNet frames that have a \emph{close match} with (i.e. are evoked by)
the synsets in the virtual documents, and we add them to it. The hierarchy of frames is also exploited to include additional, more general frames. As a result, each community is represented by a concatenation of all the retrieved synsets and frames.
Figure \ref{img:synsetframe-cluster}\footnote{\texttt{wn30:} \url{https://w3id.org/framester/wn/wn30/instances/} \texttt{frame:} \url{https://w3id.org/framester/framenet/abox/frame/}} shows the synsets and frames included in the virtual documents of the two communities from Figure \ref{img:communities}.

\noindent\textbf{Clustering.} We use K-Means \cite{macqueen1967kmeans} and the elbow method to cluster the communities' virtual documents. It is a general-purpose clustering algorithm that has been tested across different application areas and domains \cite{xu2015clusteringsurvey}.
K-Means partitions the observations into a predefined number of $k$ disjoint groups, defining, after a number of iteration, $k$ centroids, one for each cluster. \\
Clusters i.e. conceptual components, are organised as a hierarchical network. To generate relations between them we use FrameNet inheritance relations between frames.
Given a cluster $c$, we indicate with $F(c)$ the set of frames associated with $c$. Two clusters $c_1$ and $c_2$ are hierarchically related $r(c_1,c_2)$, with a weight $w$, if at least one frame $f_1\in F(c_1)$ inherits from at least one frame $f_2\in F(c_2)$.
We indicate with $max$ the maximum number of inheritance relations between frames that occur between two clusters of the network.
The weight $w$ indicates the strength of $r(c_1,c_2)$ and is computed as follows.
Given two clusters $c_1$ and $c_2$, the strength $w$ of $r(c_1,c_2)$ is the sum of the frames in $c_1$ that are subsumed under at least one frame in $c_2$, divided by $max$. The range of values for $w$ is  [0,1].

\noindent\textbf{Naming conceptual components.} To automatically assign a meaningful name to each cluster, which identifies a conceptual component, we generate (and manually check) a label from the most frequent synsets and frames that belong to it, i.e. we count how many times a same synset or frame is included in the virtual documents belonging to a cluster. We also generate a textual description for each cluster by concatenating all terms representing its communities. This description is useful to better understand the more specific concepts covered by the OODPs grouped by a cluster. For example, the communities in Figure \ref{img:synsetframe-cluster} are grouped in the same cluster named \emph{Event}: the most frequent frame within the cluster (41 times from 21 communities belonging to 13 different ontologies). The description indicates that the ontologies implementing the \emph{Event} conceptual component address cultural events, organisers, reproduction, time, etc. 

\noindent
\textbf{Catalogue generation.}
The last step of our method (cf. Figure \ref{img:pipeline}) builds a \emph{catalogue}
that connects and organises the analysed ontologies according to the extracted conceptual components and their corresponding OODPs. Each CC in the catalogue is linked to its associated OODPs within the ontologies. Therefore, the catalogue classifies the ontologies based on the conceptual components that they implement. 
We provide an HTML rendering of the catalogue\footnote{\url{https://stlab-istc-cnr.github.io/cc-and-odps-catalogue/}} included in the online package\footref{note:package}, generated from the CH corpus.
\section{Experiment and Results}
\label{sec:results}
The overall time required for producing all results with our method is about 1h15m (CH corpus) and 30m (Conf corpus) on a commodity hardware: we used a laptop (2,3 GHz Intel Core i5, 16GB of RAM).\\
\noindent \textbf{Intensional graphs.}
The average number of nodes and edges of the intensional graphs derived from the CH corpus is $\sim$165 and $\sim$217, respectively. For the Conf corpus, they are $\sim$91 and $\sim$115.
The intensional graphs preserve an average 47\% (CH corpus) and 54\% (Conf corpus) of classes and 90\% (CH corpus) and 87\% (Conf corpus) of object and datatype properties.
The loss of information about ontology classes is due to the fact that the tranformation rules defined in Rules \ref{listing:sparql-int} are biased towards ontologies with rich axiomatisation: ontologies that have poor axiomatisation are mostly affected by information-loss. Nevertheless, we remark that all superclasses and superproperties are discarded in this process, while they are all recovered when the OODPs are retrieved. \\
%
\textbf{Community detection.} We detect a total number of 1,300 communities from the CH corpus. The smallest number of communities found per ontology is 1 (RDA):
only in one case the algorithm could not split the ontology in different communities. The greatest number of communities is 363 (ArCo). 
The average number of communities per ontology is $\sim$30.
As for the Conf dataset, from 16 ontologies our algorithm detects 419 communities with an average of $\sim$26 communities per ontology. The minimum number of communities is 8, while the maximum is 83. \\
\textbf{Clustering.} We convert the virtual documents in numerical feature vectors
and apply tf-idf to discard tokens that occur too frequently. Our setting ignores terms that have a document frequency higher than 90\%. We did not fix a minimum value.
%
%
To evaluate the optimal number of clusters $k$ for our data, we used the elbow method and we run the algorithm with a fixed number of 100 (CH dataset) and 81 (Conf dataset) clusters.
Being K-Means nondeterministic, 
we set the \emph{random state} parameter to a commonly used integer value (42) in order to make our cluster assignments reproducible.\\
\noindent For the CH corpus, the average number of communities per cluster is 13, with a maximum of 111, and a minimum of 3. Each cluster contains communities that belong to an average of $\sim$4.5 different ontologies. 11 clusters group communities from the same ontology. 88 clusters group an average of $\sim$15 communities that belong to a range between 2 and 10 different ontologies. 1 cluster groups 111 communities from 26 different ontologies.\\
For the Conf corpus, the average number of communities per cluster is 5.17, with a maximum of 13, and a minimum of 1. The communities in each cluster belong to an average of $\sim$2.6 different ontologies. 25 clusters group communities from the same ontology, the remaining 56 clusters group an average of $\sim$7.2 communities that belong to a range between 2 and 8 different ontologies. \\
\textbf{Catalogue.}
Table~\ref{tab:cluster_hierarchy} gives an overview  of the number of  hierarchical relations among the clusters per level of strength (see Section \ref{sec:clustering}). 

\begin{table*}[t]
    \centering
    \caption{The number of hierarchical relations among clusters per level of strength. For each level \textit{l}, it is indicated the total, maximum and average number of relations having a strength $\ge l$.}
    \begin{tabular}{c|ccc|ccc|ccc|ccc|ccc}
    \hline
                        & \multicolumn{15}{c}{Strength levels} \\\hline
                        & \multicolumn{3}{c|}{0.0} & \multicolumn{3}{c|}{0.1} & \multicolumn{3}{c|}{0.2} & \multicolumn{3}{c|}{0.3} & \multicolumn{3}{c}{0.4} \\
                        & tot & max & avg & tot & max & avg & tot & max & avg & tot & max & avg & tot & max & avg \\ 
                        \hline\hline
                        
        \textit{CH}     & 6644 & 91 & 69.2 & 813 & 66 & 8.4 & 274 & 42 & 2.8 & 114 & 30 & 1.18 & 58 & 22 & 0.6 \\
        \textit{Conf}   & 2000 & 47 & 25.9 & 572 & 30 & 7.4 & 260 & 25 & 3.3 & 133 & 15 & 1.72 & 63 & 11 & 0.8 \\\hline
        
    \end{tabular}
    
    \label{tab:cluster_hierarchy}
\end{table*}

%
%
%
%
\section{Evaluation and Discussion}
\label{sec:discussion}
\textbf{Manual inspection of communities.} A manual inspection of the communities, focusing on both structure and labels, has been a necessary step for determining the quality and soundness of our results. 
We define four categories of communities based on their quality: bad, medium, good, ideal.
A community is \emph{bad} if it can belong to more than two CCs, it lacks a conceptual coherence and its implementation (OODP) is poorly axiomatised. 
For instance, a community from the Conf corpus includes 27 heterogeneous properties (e.g. \emph{created by} and \emph{has conflict type}) that are not involved in any restrictions e.g. range or domain. 
A CH community involves unrelated properties having the same domain and \texttt{xsd:string} as range. In these cases, the topology could not support the identification of significant modules, while the vocabulary highlights the presence of various conceptual areas\footnote{Detecting bad communities may be a useful tool for evaluating the quality of an ontology.}.
A community has \emph{medium} quality if it can belong to two CCs.
A community is \emph{good} if it can belong only to one CC but includes max 20\% of intruders (incoherent entities). \emph{Ideal} communities have less than 20\% of intruders.\\
About 8\% of communities in both the CH and Conf ontologies are \emph{bad}.
$\sim$7\% (CH) and $\sim$3\% (Conf) have \emph{medium} quality, $\sim$17\% (CH) and $\sim$5\% (Conf) are \emph{good}, while 
the majority of the communities (CH: $\sim$67\%, Conf: $\sim$84\%) have an \emph{ideal} level of semantic coherence e.g. see the two implementations of the CC \emph{Event} of Figure \ref{img:synsetframe-cluster}. \\
Let us take two additional examples from both corpora. A community from the Conf dataset (cmt-2 ontology) identifies an OODP for \emph{being a member of a conference}: it includes the two 
inverse binary relations and the concepts \emph{conference} and \emph{conference member}, which are their domain/range. A CH community from CIDOC CRM implements an OODP for capturing that an object changed its ownership: it includes the concept \texttt{crm:E8\_Acquisition}, and the predicates representing the physical entity and the actors that acquired and surrendered the title over it. By inspecting the OODP, we found that all properties in this fragment have domain and range, but inverse object properties are not asserted.\\
\textbf{Clustering: similarity.} 
For assessing the quality of the clusters we computed a pairwise similarity among them.
Specifically,  we adopted the Overlap Coefficient\footnote{\url{https://en.wikipedia.org/wiki/Overlap_coefficient}} (commonly used in data mining techniques) for measuring the overlap between the sets of synsets and frames of two clusters.
This score indicates how similar two clusters are and its values ranges from 0.0 (dissimilar) to 1.0 (similar).
We observed that, on average, the clusters of both corpora score very low (0.20 for CH and 0.17 for Conf) thus indicating a good quality of the clusters.

\noindent \textbf{Clustering: manual inspection.} 
The clusters that have been detected in both datasets identify a wide range of different conceptual components, with different levels of abstraction. General components such as \emph{Event}, \emph{Categorization}, \emph{Membership}, \emph{Intentionally act} emerge from both corpora. This finding can support interoperability between ontologies addressing different domains.
Other components are more specific to the domain: e.g. \emph{Performing arts}, \emph{measurement} and \emph{attribution} from the CH corpus; \emph{Submitting documents}, \emph{Respond to proposal}, \emph{Award} from the Conf corpus. By inspecting a CC, it is possible to compare implementations from different ontologies and choose the one that best fits our requirements: for example, the CIDOC CRM and ArCo implementations of the CC \emph{Acquisition} overlap only partially: ArCo also addresses the acquisition place and time, while it does not model the new owner as in CIDOC.
%
%
\\
In both corpora, some clusters could be either split or merged. If no frames/synsets clearly emerge for a cluster this may indicate that it groups different conceptual components. 
The emergence of the same frame(s) as the most frequent in different components may indicate that they could be merged, or that they are a specialisation of the same conceptual component: this can be clarified by looking at less frequent frames and at their hierarchical relations. 
\\\noindent
\textbf{Evaluation against an ontology engineering task.} We evaluate our method by also analysing our results in the context of an ontology matching task. While it is an indirect evaluation, we believe it is informative of the quality of our approach. 
%
The hypothesis for this evaluation is that given a pair of entities that shall be aligned (through subsumption or equivalence), these entities should belong to either the same cluster or two related clusters.
\noindent Intuitively, since a cluster groups semantically close OODPs from different ontologies, an agent (human or artificial) performing ontology alignment on a corpus of ontology, can look within a same cluster or follow strong hierarchical relations between clusters to identify entities that shall be aligned. The question is whether a good number of these alignments can be identified with this approach.
%
We use three sets of alignments to compare our results: (i) a set AA of 224 asserted (curated) alignments (1 equivalence and 223 subsumptions) from the CH corpus; (ii) a set AML of 237 alignments (all equivalences) generated by AgreementMakerLight~\cite{Faria2013} (the best tool in most of the OAEI 2020 tracks), for all pairs of ontologies in the CH corpus; (iii) a dataset CA of 224 alignments on the Conf corpus (all equivalences) used as gold standard in the OAEI 2020 conference track\footnote{\url{http://oaei.ontologymatching.org/2019/conference/data/reference-alignment.zip}}.
\\ \noindent The AML dataset associates a confidence score $cs$ with each alignment, while for the CA and AA sets we assume a $cs$ = 1 for all alignments. 
We introduce $AML_{.90} \subseteq AML$ and $AML_{.99}\subseteq AML$ which are the sets containing the alignments having a confidence score $\ge 0.9$ and $\ge 0.99$, respectively.
%
%
%
To measure the quality of our results, we assume that given a set of alignments $A$, the pairs of entities belonging to $A$ are assigned to a same cluster or to related clusters, with the same $cs$ provided for that alignment. For example, if a pair $(e1,e2)$ belongs to AML with a confidence score $cs = 0.98$, then we assume that AML would assign $(e1,e2)$ to the same cluster or to two related clusters with $cs = 0.98$. Finally, we introduce the sets \textit{D}, \textit{I}, \textit{H} and \textit{E} to interpret  the results of our method.
Given a set of entity pairs \textit{D} from the alignment in AA, CA or AML, we define: (i) the set $I \subseteq D$ as the set of  entity pairs in $D$, that belong to same clusters; (ii) $H \subseteq D$ as the set of entity pairs that belong to hierarchically related clusters; and (iii) $E := I \cup H$. With $H_{n}$ (similarly $E_{n}$) we indicate the set of entity pairs that belong to two clusters related with strength $l \ge n$\footnote{We remind that n = [0,1] indicates the strength of the hierarchical relation between two clusters.}.

We propose the measure \textit{corr} (cf. Formula \ref{def:corr}) to compute the correlation between the alignment sets and the results of our method. Given two sets of entity pairs \textit{A} and \textit{B}, each pair assigned with a confidence score $cs(e_{i},e_{j})$, we define \textit{corr(A,B)} as the sum of all $cs$ of the alignments in \textit{A} divided by the sum of all $cs$ in \textit{B}, that is:
\begin{align}
\label{def:corr}
corr (A,B) = \frac{\sum\limits_{(e_i,e_j) \in A} cs((e_i, e_j))}{\sum\limits_{(e_i,e_j) \in B} cs((e_i, e_j))}
\end{align}
The $cs$ associated with the alignments of AA and CA is 1.0. The correlation ranges from 0.0 (no correlation) to 1.0 (strong correlation). The entity pairs from our method inherit the $cs$ value from the comparing set.
Intuitively, $corr$ computes the ratio between the pairs that should be aligned and the pairs that belong to same or related clusters. Table~\ref{tab:alignment_analysis} reports the value of $corr$ computed for comparing \textit{AA, CA, AML,} $AML_{.90}$ and $AML_{.99}$ (the testing sets) with the sets $I$, $E$, $E_{n}$, $E_{0.2}$, $E_{0.3}$, $E_{0.4}$.

\begin{table}[t]
    \centering
     \caption{Correlation between reference alignments (\textit{AA, CA, AML,} $AML_{.90}$ and $AML_{.99}$) with the sets   $I$, $E$, $E_{0.1}$, $E_{0.2}$, $E_{0.3}$, $E_{0.4}$.}
    \begin{tabular}{ccccccc}
        \hline
        \textit{Alignments} & $I$ & $E$ & $E_{0.1}$ & $E_{0.2}$ & $E_{0.3}$ & $E_{0.4}$ \\\hline\hline
        
        \textit{AA}  & .21 & .99 & .64 & .46 & .34 & .32 \\
        \textit{AML} & .47 & .99 & .77 & .64 & .58 & .56 \\
        $AML_{.90}$  & .46 & .99 & .77 & .64 & .57 & .55 \\
        $AML_{.99}$  & .51 & 1.0 & .75 & .63 & .60 & .57  \\
        CA           & .27 & .76 & .51 & .43 & .36 & .35 \\\hline
    \end{tabular}
    \label{tab:alignment_analysis}
\end{table}
\noindent
\textbf{Discussion.}
Almost all CH entity pairs aligned in the testing sets ($corr \ge .99$) can be found either in same clusters or in two related clusters, a lower number for $CA$ pairs ($corr = .76$) (see column $E$ of Table \ref{tab:alignment_analysis}): all hierarchical relations between clusters are to be inspected in the worst case (69.2/CH and 25.9/Conf, on average per cluster). The dimension of the task may sound inconvenient for manual inspection, nevertheless we remark that an entity-to-entity analysis of the ontologies in the CH/Conf corpus would require the inspection of 43/16 ontologies and, in the worst case, of 11839/1565 predicates.
An artificial agent e.g. an ontology alignment algorithm, may use clusters and their relations to inform a strategy for ranking candidate pairs in a corpus (at the moment, ontology alignment tools works with two ontologies at a time). By setting a threshold for $l$, i.e. discarding weaker hierarchical relations, the value of $corr$ decreases, but it holds reasonably good for the CH corpus until up to $l = 0.3$ (with only 1.18 average relations per cluster). With $l = 0.4$, it is possible to find up to 57\% of the most precise alignments ($AML_{.99}$) by looking to entities belonging to same clusters. As for $AA$, the performance are the the worst in our experiment e.g. for column $I$. To better understand this result we run AgreementMakerLight on the CH corpus and compare its results against AA (which are curated/asserted alignments). We report that only 1.5\% of the alignments are identified. Our approach does not identify alignments, hence we cannot claim to perform better than AgreementMakerLight, however we speculate that this result (cf. Table \ref{tab:alignment_analysis}), as compared to this extremely low performance, supports our hypothesis that clusters and their relations may be used to improve the performance of alignment algorithms. 
\section{Related work}
\label{sec:related}
\textbf{Ontology selection and understanding.} 
Catalogues of ontologies (e.g. \href{http://purl.org/vocab/}{vocab.org})
and ODPs (e.g. \href{http://ontologydesignpatterns.org/}{ontologydesignpatterns.org})
and semantic search engines (e.g. \href{http://prefix.cc/}{prefix.cc}) are meant to support ontology selection for reuse. Users can browse ontology terms, but comparing multiple ontologies is not supported. None of them support ODP-based browsing or filtering.
Most ontology summarisation approaches, e.g. those cited in \cite{summarization-survey2018-arxiv,cebiric2018summarizing}, look for the most informative concepts/nodes using centrality measures, PageRank and the like. Or extract relevant subgraphs to support query-testing for validating requirements against available data. To the best of our knowledge, this is the first method using an ODP-based approach, which provides an ontology designer with relevant small ontology fragments to reuse based on specific modelling problems. \\
\textbf{Ontology partitioning.} Modularisation approaches e.g. \cite{amato2015partitioning, dAquin2009modularization, ghafourian2013graph}
work on single ontologies and return non-overlapping, consistent modules, that combined together form the original ontology \cite{dAquin2009modularization}. They mainly focus on logical and structural modularisation, and no additional insight about the modules is provided, while each cluster of OODPs (CC) is given a name, description and images. \\
%
\textbf{Complex ontology matching.} Complex ontology matching is the process of generating complex alignments, containing at least one entity 
on which a constructor or a transformation function is applied \cite{thieblin2019}. An ODP-based approach to this task is proposed by \cite{fillottrani2017}, which also provides a formalisation of the common structure of two (or more) aligned patterns (a potential logical characterisation of CCs). Our method may be the basis to novel approaches/implementations to address this research task. \\
\textbf{Patterns discovery.} Ontology patterns discovery consists in finding frequent repeating structures. \cite{mikroyannidi2011inspecting} clusters repetitive structures of axioms, and then generalise them, while we start from detecting dense communities in ontologies, that are then clustered based on their vocabulary. The method by
\cite{lawrynowicz2018discovery} 
proposes a tree-mining method, that transforms ontology axioms in a tree shape and uses association analysis to mine co-occurring axiom patterns.
However, this method does not take into account inferences, nor the influence of the vocabulary.
\section{Conclusion and ongoing work}
\label{sec:conclusion}
Our method implements a non-extractive technique, to support understanding and comparison of multiple ontologies. It combines community detection, word sense disambiguation, frame detection and clustering to automatically generate a catalogue of conceptual components and observed ontology design patterns, starting from a corpus of ontologies. The catalogue classifies the ontologies according to the conceptual components they implement. We show its potential by testing and evaluating it on two corpora in the CH and Conference domains. While our experiments show satisfying results, they also point out improvements and research challenges. Class expressions shall be included in the intensional graph. Studying heuristics for refining CCs (split/merge), improving their naming/description and ranking them in the catalogue is an immediate next step. A user-based evaluation of the catalogue is also in our plans, however it is not as easy as evaluating key concepts, it requires involving experts in pattern-based ontology design. Automatically linking observed patterns to ODP catalogues, and managing a possible synchronous evolution of both resources, is a research challenge worth a huge impact on interoperability on the Semantic Web.

\emph{Acknowledgements.} This work has been enabled by the H2020 Project \emph{Polifonia: a digital harmoniser for musical heritage knowledge} funded by the European Commission Grant number 101004746.
%


\printbibliography

\end{document}